\documentclass[sigconf]{acmart}
\usepackage{booktabs} 

\acmDOI{http://dx.doi.org/10.1145/3097983.3098005}
\acmISBN{ISBN 978-1-4503-4887-4/17/08}

\acmConference[]{}{}{} 
\acmYear{2017}
\copyrightyear{2017}

\setcopyright{none}
\begin{document}

\title{Deep Choice Model Using Pointer Networks for Airline Itinerary Prediction}

\author{Alejandro Mottini}
\affiliation{%
  \institution{Innovation and Research Division, Amadeus SAS}
  \city{Sophia Antipolis} 
  \country{France}
}
\email{alejandro.mottinidoliveira@amadeus.com} 

\author{Rodrigo Acuna-Agost}
\affiliation{%
  \institution{Innovation and Research Division, Amadeus SAS}
  \city{Sophia Antipolis} 
  \country{France}
}
\email{rodrigo.acunaagost@amadeus.com}


\begin{abstract}
Travel providers such as  airlines and on-line travel agents are becoming more and more interested in understanding how passengers choose among alternative itineraries when searching for flights. This knowledge helps them better display and adapt their offer, taking into account market conditions and customer needs. Some common applications are not only filtering and sorting alternatives, but also changing certain attributes in real-time (e.g., changing the price). In this paper, we concentrate with the problem of modeling air passenger choices of flight itineraries. This problem  has historically been tackled using classical Discrete Choice Modelling techniques. Traditional statistical approaches, in particular the Multinomial Logit model (MNL), is widely used in industrial applications due to its simplicity and general good performance.  However, MNL models present several shortcomings and assumptions that might not hold in real applications. To overcome these difficulties, we present a new choice model based on Pointer Networks. Given an input sequence, this type of deep neural architecture combines Recurrent Neural Networks with the Attention Mechanism to learn the conditional probability of an output whose values correspond to positions in an input sequence. Therefore, given a sequence of different alternatives presented to a customer, the model can learn to point to the one most likely to be chosen by the customer. The proposed method was evaluated on a real dataset that combines on-line user search logs and airline flight bookings. Experimental results show that the proposed model outperforms the traditional MNL model on several metrics.

\end{abstract}


\maketitle

\section{Introduction}

Understanding passenger behaviour and their itinerary preferences is an important problem in the travel industry.
Different players in the sector have diverse needs but could all benefit from accurate itinerary choice prediction.  This can be used, for example, to better estimate demand and market shares in the context of dynamic markets.

In this work we concentrate in particular on the airline itinerary choice prediction problem. Consider for example that a customer is searching for flights from New York to London departing next Tuesday and coming back on Saturday. This search request is processed by a travel provider (e.g., airlines or on-line travel agents). The provider could propose up to 200 different alternatives, also called itineraries, to the customer. They are displayed in one or several pages in a predefined order (e.g., by price). Given the offer, the customer considers  different attributes of the alternatives to make the decision, such as the number of stops, total trip duration, and notably price. Therefore, the key relevant question to travel providers is: ``which alternative is most likely going to be selected by the customer?"

Predicting the user's choice has many direct applications, such as filtering alternatives (e.g., showing only the top 20), sorting them differently or even changing some attributes in real-time (e.g., adding or removing some ancillary services). Moreover, these models can be used to perform revenue management and price optimization \cite{dyn_price}. This is beneficial for all involved parties: travel providers can increase their revenue and conversion rates while passengers can  find the most relevant flights covering their needs faster. 

Historically, these kinds of problems have been  tackled using Discrete Choice Modeling (CM). CM  is an important area of research in diverse fields such as economics \cite{eco_research}, marketing \cite{cm-marketing}, and artificial intelligence  \cite{ai_research}. Moreover, this type of model is widely used in many industries such as retail \cite{retail_research} and transportation \cite{transp_research}. 
The CM framework was originally proposed by Nobel prize winner Daniel McFadden \cite{mnl}, and has been the basis for all the subsequent research in the field.  In this seminal work, McFadden introduces the Multinomial Logit model (MNL). It is the most widely used model in industrial applications due to its simplicity, good performance and ease of interpretation. In particular, it is the most popular approach for air travel itinerary choice prediction \cite{airlineApp1,airlineApp2,airlineApp3}.  


In spite of these advantages, MNL models present some weaknesses. First of all, the model only considers a linear combination of the input features, which can limit its predictive capability. Secondly, the model  suffers from the Independence of Irrelevant Alternatives (IIA) property \cite{mnl_iia}, which states that if choice 1 is preferred to choice 2 out of the choice set ${1,2}$, introducing a third option 3 (thus expanding the choice set to {1,2,3})  cannot make 2 preferable to 1. Finally, the MNL formulation cannot take the order of the alternatives into account. 

These shortcomings might be overly restrictive and cause inaccurate results for some applications \cite{ref_mnl_probelms}. In particular, real industrial applications require different models for distinct markets. In the case of air travel itinerary prediction, this involves estimating models at a city-pair level \cite{airlineApp2} and/or customer demographic segment \cite{airlineApp3, paperRodrigo}.

To deal with these limitations, in this work we propose a new Deep Choice Model (DCM) based on Pointer Networks (Ptr-Net) \cite{ptr-net}.  This type of model combines  Recurrent Neural Networks (RNN) with the Attention Mechanism \cite{attMec} in an encoder-decoder architecture.  Ptr-Net specifically targets problems where the outputs are discrete and correspond to positions in the input. Given an input sequence, the model learns the conditional probability of an output whose values are positions in an input sequence. Thus,  the output distribution over the dictionary of input choices represents the estimated probability of choice for all alternatives. This type of model has recently been applied to different problems \cite{ptr_net_aplication1, ptr_net_aplication2}.  In particular, \cite{ptr_net_aplication1} proposes a new  generative model for programming code generation that combines pointer networks to copy words from the recent input context and a character-level softmax classifier to produce other tokens in the vocabulary.


We would like to emphasize that our problem  is not a standard labelling or detection task  because we have prior knowledge of the number of 
results in each class (only one chosen itinerary per user). In addition, the alternatives are not the same  for all user sessions. Two different users can both choose their respective "alternative 1" in their sessions, but those alternatives can represent completely unrelated itineraries. Furthermore, it should be noted that there are two other fields related to our problem, but that are not directly applicable: learning to rank \cite{ranking}  and  recommender systems \cite{recsys}. Ranking methods are not directly applicable since we only have one positive case (the choice) and  all others alternatives are negative cases. There are no ``intermediate choices". On the other hand, given that each user session is anonymous and there is no user history in our dataset, classical recommender system algorithms cannot be directly used for this application.

We validate the effectiveness of our deep choice model on a dataset combining real  on-line user search logs and airline flight bookings. Experimental results show that the proposed model  outperforms the traditional MNL method as well as a Gradient boosting tree based model on different metrics. In particular, the alternative with the maximum estimated probability can be compared to the real choice to calculate the top-1 and top-N accuracy of the model, along with other business related metrics.

The main contributions of our paper are twofold. First, we propose a novel approach to model choices based on Pointer Networks, which solves some of the shortcomings of the MNL model. To the best of our knowledge, this is the first time  this neural network  architecture has been used to model discrete choice problems, in a field that is clearly dominated by MNL models. The analogy between Discrete Choice Modelling and Pointer Networks is simple but powerful: the input sequence correspond to the choice set and the output is a pointer to the most probable alternative. 
Secondly, our approach obtains better prediction results than other tested methods, and presents practical advantages when it comes to industrial implementations.  Our model allows us to work with numerical and categorical features without feature engineering, and at the same time be trained with heterogeneous data. This is a clear advantage compared to MNL models, where data usually needs to be segmented (e.g., at city-pair level) before estimating the models. Our experiments on multi-market data show better prediction capabilities for our proposed approach compared to the traditional models used in the industry, which simplifies the development, storage and maintenance of industrial applications.

\section{Related Work}
\label{sec:rwork}

As mentioned before, Discrete Choice Modeling is a well-studied problem in various fields of research. Nevertheless, most research has been so far concentrated on MNL and its variants. In particular, richer models such as Nested logit model \cite{nested} and the hierarchical MNL \cite{h_mnl} have been studied in the literature and can capture more complex choice behaviours. Moreover, these and other extensions avoid the IIA property. As a shortcoming, this added complexity results in more complex optimization problems. For example, Davis et al. \cite{np_nested} have shown that the optimization of the  Nested logit model is in general a NP-hard problem.

Moreover, Blanchet et al. \cite{markov} propose a Markov chain based choice model, where the substitution from one product to another is modeled as a state transition of a Markov chain. The chain's parameters are estimated with a data-driven procedure.

In addition, there is a family of methods which is more concerned with correctly simulating the human choice making process \cite{human}.
Research in this area has revealed different phenomena influencing human choice (e.g., the similarity effect, the compromise effect, and the attraction
effect) \cite{human_choice} and tries to define models able to capture them from choice data.

More related to our work, \cite{aan-mnl} proposes to modify the MNL model by reformulating the utility equation using a feed-forward multilayer neural network. The model is referred to as AAN-MNL and is able to consider non-linear effects of the features. Finally, \cite{dcm1} describes a model based on Restricted Boltzmann Machines. The model could not handle choice's features, which significantly limited  its applicability. The model was recently extended by the authors in \cite{dcm2} to incorporate features from images extracted through deep learning as input to the original model.


\section{Discrete Choice Model}
\label{sec:cmodel}



Discrete choice models have been used by researchers and practitioners in many industries to predict choices 
between two or more discrete alternatives. All discrete choice models share the following three basic components: a decision maker, a choice set, and the choice. The collection of alternatives presented to a decision maker is sometimes referred as a session.

Faced with a set of finite choices, the decision maker (user) must choose one of them. This choice is usually modeled as a binary variable. It is assumed that the user takes a rational decision based on his tastes and needs by considering the attributes of the proposed alternatives.

Moreover, the choice set needs to verify three basic conditions: a) mutually exclusive, b) exhaustive, and c) be composed of a finite number of alternatives. Condition (c) is a key aspect to be considered when selecting between a regression analysis or a discrete choice model.  Given this three elements, the objective is to learn the choice model of how users choose among products.

\subsection{Multinomial Logit Model}

The MNL framework is derived under the assumption that a decision maker chooses the alternative that maximizes the utility he receives from it.  
Formally, a decision maker $i \in I$ chooses between $J$ alternatives. He would obtain a certain utility $U_{i,j}$ from each alternative $j\in J$, and choose alternative $\hat{j}$ if and only if:
\begin{equation}
U_{i,\hat{j}} > U_{i,j} ; \forall j \neq \hat{j} 
\end{equation}

In practice, the utility function is unknown and not observable.  However, we can determine some features of the alternatives as faced by the decision maker, denoted as $x_{i,j} \forall j$. In addition, we might have attributes associated to each decision maker, denoted $s_i$. Based on these variables, we can define a model that relates the observed features to the unknown decision maker's utility:
\begin{equation}
V_{i,j}=V(x_{i,j},s_i)
\end{equation}
where $V_{i,j}$ is referred to as representative utility and is generally a linear combination of the features. For example, if an airline is trying to predict which itinerary a user will choose, a very simple model could be:
\begin{equation}
 V_{i,j} = a*price_{i,j} + b*tripDuration_{i,j}
\end{equation}
where $a,b$ are parameters of the model to be estimated. In general, the model is not perfect and $V_{i,j} \neq U_{i,j}$. The relationship between both quantities can be expressed as:
\begin{equation}
U_{i,j} = V_{i,j} + \varepsilon_{i,j}
\end{equation}
where $ \varepsilon_{i,j}$ is a random term that encapsulates all the factors that impact the utility but are not considered in $V_{i,j}$.

We can express the probability that decision maker $i$ chooses alternative $\hat{j}$ as:
\begin{equation}
P_{i,\hat{j}} = P(U_{i,\hat{j} } > U_{i,j} ; \forall j \neq \hat{j} )
\end{equation}

In \cite{mnl} the author shows that if $ \varepsilon_{i,j}$ are i.i.d  Gumbel random variables, the MNL model has the following key property:
\begin{equation}
P_{i,j}= \frac{\exp(V_{i,j}) } { \sum_{k=1}^J \exp(V_{i,j}) }
\end{equation}

Finally, the model is optimized using maximum likelihood estimation:
\begin{equation}
\hat{\theta} = \underset{\theta}{\arg\max} \sum_{i \in I} \sum_{j \in J}   y_{i,j} \ln(  P_{i,j} )
\end{equation}
where $y_{i,j}$ is a binary indicator of whether decision maker $i$ is associated with the choice $j$. Different optimization algorithms can be used to numerically find a local optima of this log likelihood function.

\section{Deep Choice Model}
\label{sec:deepCM}

In this section we will start by describing the Pointer Network framework and previous architectures on which it is based. We will then detail the proposed deep choice model.

\subsection{Pointer Network}

Pointer Networks (Ptr-Net) were originally proposed by Vinyals et al. \cite{ptr-net}. These neural architectures  combine the popular sequence-to-sequence (Seq2seq) learning framework  \cite{seq2seq} with a modified Attention Mechanism \cite{attMec}.


Seq2seq models have two main components: an encoder and a decoder network. The encoder  maps a variable length input sequence into a fixed-dimensional vector representation, while the decoder transforms this vector to a variable length output sequence.  

Formally, given an input sequence $X=(x_1,...,x_n)$ of $n$ vectors and $Y=(y_1,...,y_m)$ its corresponding output sequence whose length can be different, the Seq2seq models calculates the following conditional probability:

\begin{equation}
p(Y|X)=\prod_{i=1}^m p(y_i|y_1,...,y_{i-1},X) 
\end{equation}

If we model both the encoder and the decoder with Recurrent Neural Networks (RNN) of hidden states $(e_1,...,e_n)$ and $(d_1,...,d_m)$ respectively, each conditional probability can be expressed as:
\begin{equation}
\begin{split}
p(y_i|y_1,...,y_{i-1},X) = g( y_{i-1}, d_i, c) \\
c = q({e_1,...,e_n}) \\
e_j=f(x_j, e_{j-1})
\end{split}
\end{equation}
where $q({e_1,...,e_n})=e_n$ in the simplest case, and $f, g$ are transformation functions associated to the type of RNN unit being used. In particular, \cite{seq2seq} uses a Long Short Term Memory (LSTM) cell \cite{lstm}, although other types could potentially be used. 

The encoder is fed sequence $X$, one element at a time until the end of the sequence is reached. The end of the sequence is marked by a special end-of-sequence symbol. The model then switches to decoder mode, where the elements of the output sequence are generated one at a time until the end-of-sequence symbol is generated. At this moment, the process ends. Note that unlike the model presented in Section \ref{sec:cmodel}, this type of model makes no statistical independence assumptions.

By connecting the encoder and decoder with an attention module \cite{attMec}, the decoder can consult the entire sequence of the encoder's states, instead of only the final one. This allows the decoder to focus on different regions of the source sequence during the decoding process, which improves results significantly . 

In this new model, $c$ is no longer constant and equal to the last encoder state. Therefore, each conditional probability is now defined as:
\begin{equation}
\begin{split}
p(y_i|y_1,...,y_{i-1},X) = g( y_{i-1}, d_i, c_i) \\
d_i= h(d_{i-1}, y_{i-1}, c_i)
\end{split}
\end{equation}

\begin{figure}
\centering
\includegraphics[scale=0.6]{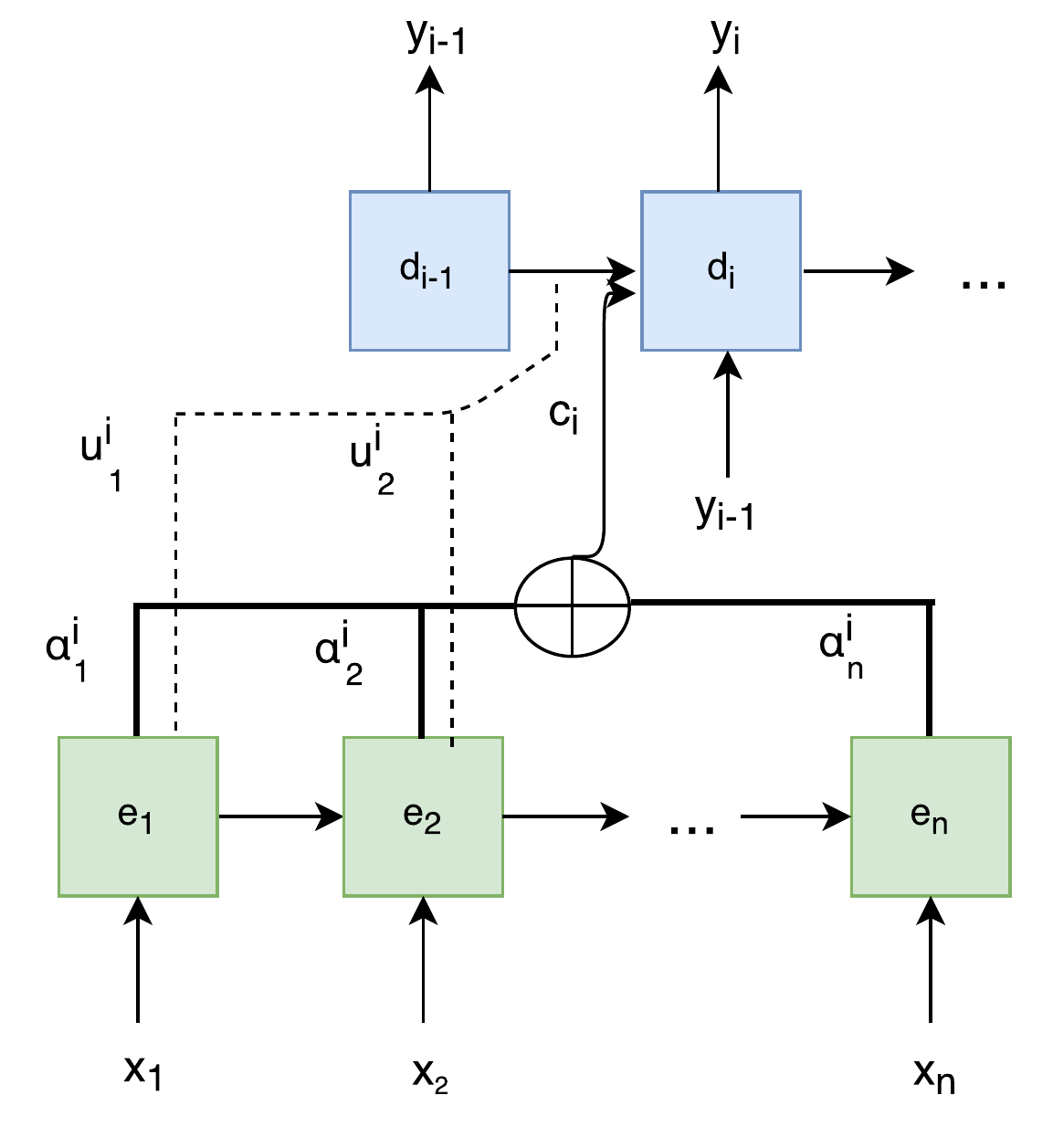}
\caption{Sequence to sequence model with attention mechanism (encoder in green, decoder in blue). }
\label{fig:att}
\end{figure}

The new $c_i$ vector is computes as follows:
\begin{equation}
c_i=\sum_{j=1}^{n} \alpha_j^i e_j
\end{equation}
where the weights $ \alpha_j^i$ are defined as:
\begin{equation}
\begin{split}
\alpha_j^i = \frac{\exp (u_j^i)}{\sum_{k=1}^{n}\exp(u_k^i)} \\
u_j^i= a(d_{i-1}, e_j)
\end{split}
\end{equation}
where $a$ is  modeled as a feed-forward neural network (jointly trained with the rest of the system) and the softmax function is used to normalize vector $u_j^i$. This normalized vector is referred to as the attention mask (or alignment vector) over the inputs.  The process is summarized in Figure \ref{fig:att}.

Although it has been shown that the additional information available to the decoder significantly improves the results of seq2seq, this does not solve the fact that the output dictionary depends on the length of the input sequence. 

Ptr-Net achieves this by adapting the attention mechanism to create pointers to elements in the input sequence. The following modification to the attention model was proposed: 
\begin{equation}
\begin{split}
u_j^i = v^T \tanh(W_1e_j+W_2d_i) \\
p(y_i|y_1,...,y_{i-1},X)=  \frac{\exp (u_j^i)}{\sum_{k=1}^{n}\exp(u_k^i)} \\
\end{split}
\label{eq:ptr}
\end{equation}
where $v, W_1, W_2$ are learnable parameters. Softmax normalizes vector $u$ to be an output distribution over the dictionary of inputs. It should be noted that unlike the  standard attention mechanism,  the Ptr-Net model does not use the encoder states to propagate extra information to the decoder, but instead uses $u_i^j$ as pointers to the input sequence elements.

\subsection{Deep Choice Model Using Pointer Networks}

\begin{figure}
\centering
\includegraphics[scale=0.5]{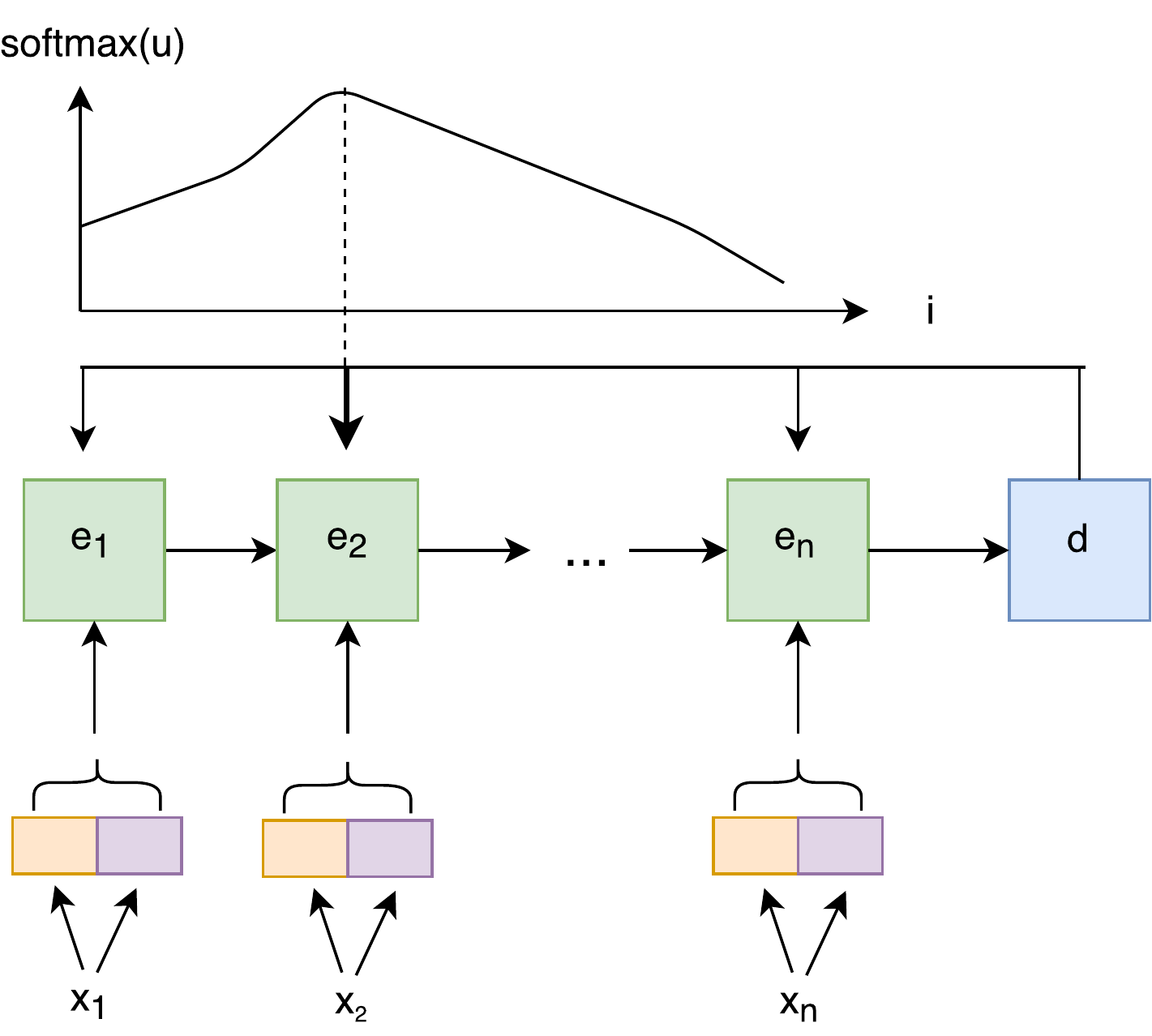}
\caption{Proposed deep choice model using pointer network (encoder in green, decoder in blue, normalization in orange and embedding in purple). }
\label{fig:pointer}
\end{figure}

The overall structure of our system is illustrated in Figure  \ref{fig:pointer}. As discussed in the previous section, it is made up of an encoder-decoder network that uses the modified pointer-network attention mechanism. However, we propose some modifications to the original Ptr-Net algorithm.

In the original Ptr-Net formulation, the authors apply the method to applications such as sorting number sequences and calculating the convex hull of a series of points in space. These problems require a RNN decoder to produce an output sequence that proposes a candidate element one step at a time until a ``stop position" is predicted. For example, if the model sorts lists of 10 random numbers, the decoding process will start by inputting a $<$GO$>$ symbol to the decoder. The decoder will output a $u$ vector (see Equation \ref{eq:ptr}) that will point to the location of the list's element most likely to be the first element of the sorted list. This first prediction will be used as the next input of the decoder, which will produce the second element of the sorted list, and so on. The generation process will end when the decoder points to a special $<$EOS$>$ position. This special position requires the model to have $C+1$ output classes, where $C$ is the number of possible choices.

In our application, we do not need to produce an output sequence. We are able to sort the alternatives (and  determine the most likely choice) by simply using the  $u$ vector from the first decoding step. Therefore, we remove the additional $<$EOS$>$ position from the model. A RNN decoder is also no longer needed.


In addition, the formulation used in Equation \ref{eq:ptr} is just one possible way of comparing the decoder vector and the encoder states. Instead, we propose to use a different method originally proposed in \cite{att_mod_ptr}. Thus, the final equations are:
\begin{equation}
\begin{split}
d=  \tanh(W_2e_n+b) \\
u_j =d^T W_1e_j\\
p(y_j|X)=  \frac{\exp (u_j)}{\sum_{k=1}^{n}\exp(u_k)} \\
\end{split}
\label{eq:ptr_mod}
\end{equation}
where $d$ no longer depends on $i$ and the alignment vector between the decoder vector and the encoder states is computed using a simpler equation that results in a better performance for our application. Finally, $p(y_j|X)$ is used to sort the alternatives presented to the user and to choose the most likely.



On the other hand, our encoder's structure remains unchanged with respect to the original  Ptr-Net method, but an additional feature pre-processing layer needs to be added. The encoder takes as input the itinerary's features (see Section \ref{sec:validation}). Numerical features (such as ticket price) are normalized to the [0,1] interval to remove the network's sensitivity to scale. In addition, embeddings are used to map categorical features to vectors. Embeddings work as lookup tables of $N$ rows and $D$ columns, where each row corresponds to  an element in the input vocabulary, and each column a latent dimension. The input vocabularies (one per categorical feature) are computed before training. All rows containing out-of-vocabulary values are assigned a special symbol $<$UNK$>$.  The dimensionality of an embedding matrix $D$ associated to feature $f$ is set such that:
\begin{equation}
D=k*\log(\#f)
\end{equation}
where $\#f$ is the cardinality of feature $f$ and $k$ a hyper parameter of the model that is usually in the $[1,10]$ interval. Each feature has a separate embedding matrix, which is initialized randomly and learned jointly with all other model parameters through back-propagation. This process produces dense representations of the features, which are more suitable  for neural networks than the sparse vectors produced by the classical one-hot encoding method. All pre-processed features are concatenated into an array and input into the encoder. The encoder will read the alternatives per session one step at a time.

Finally, to be able to handle user sessions with different number of alternatives in batch mode, a special PAD itinerary is included in sessions containing less than the maximum number of alternatives in the dataset. 




\section{Validation}
\label{sec:validation}


As part of this study, we have access to anonymized booking data from different airlines
collected by the Global Distribution System (GDS) Amadeus. GDS is a network operated by a vendor that enables automated
transactions between airlines and travel agencies. 

In the travel industry,  whenever a travel reservation is made, a Personal Name Record (PNR) is created \cite{pnr}.
It can be generated by airlines or other travel agents. PNR records will always contain the travel itinerary of the traveller, and may also include other data elements such as personal information (name, gender, age, etc), payment information (currency, total price, etc) and additional ancillary services sold with the ticket (such as extra baggage and hotel reservation). 

An anonymized subset of PNRs is stored in a dataset called MIDT (Marketing Information Data Tapes). As one of the world's GDS, Amadeus MIDT has detailed reservation data on all air bookings made by partner Travel Agencies on all participating carriers, which includes approximately 420 airlines  and activity reported from over 93000 Travel Agency locations.

However, having only access to booking data is insufficient to fully understand choice behaviour.
Therefore,  we have also used a large data source coming from search logs (i.e., what people are searching/requesting the GDS).
These search logs contain not only the travel requests (e.g., origin, destination, dates), but also
complete information about the market context. In other words, we have access to the travel
alternatives that the customer saw in his screen when booking, which includes among others: different airlines,
flight numbers, time of flights, and prices.

By matching both datasets, the result contains a set of alternatives presented to each user and their corresponding choice.
There is exactly one booking per user session (set of alternatives). Moreover, there can be between 1 and 50 possible alternatives per session, which are sorted by increasing price. This is the way most flight search engines present their results to users.

\begin{figure}
\centering
\includegraphics[scale=0.41]{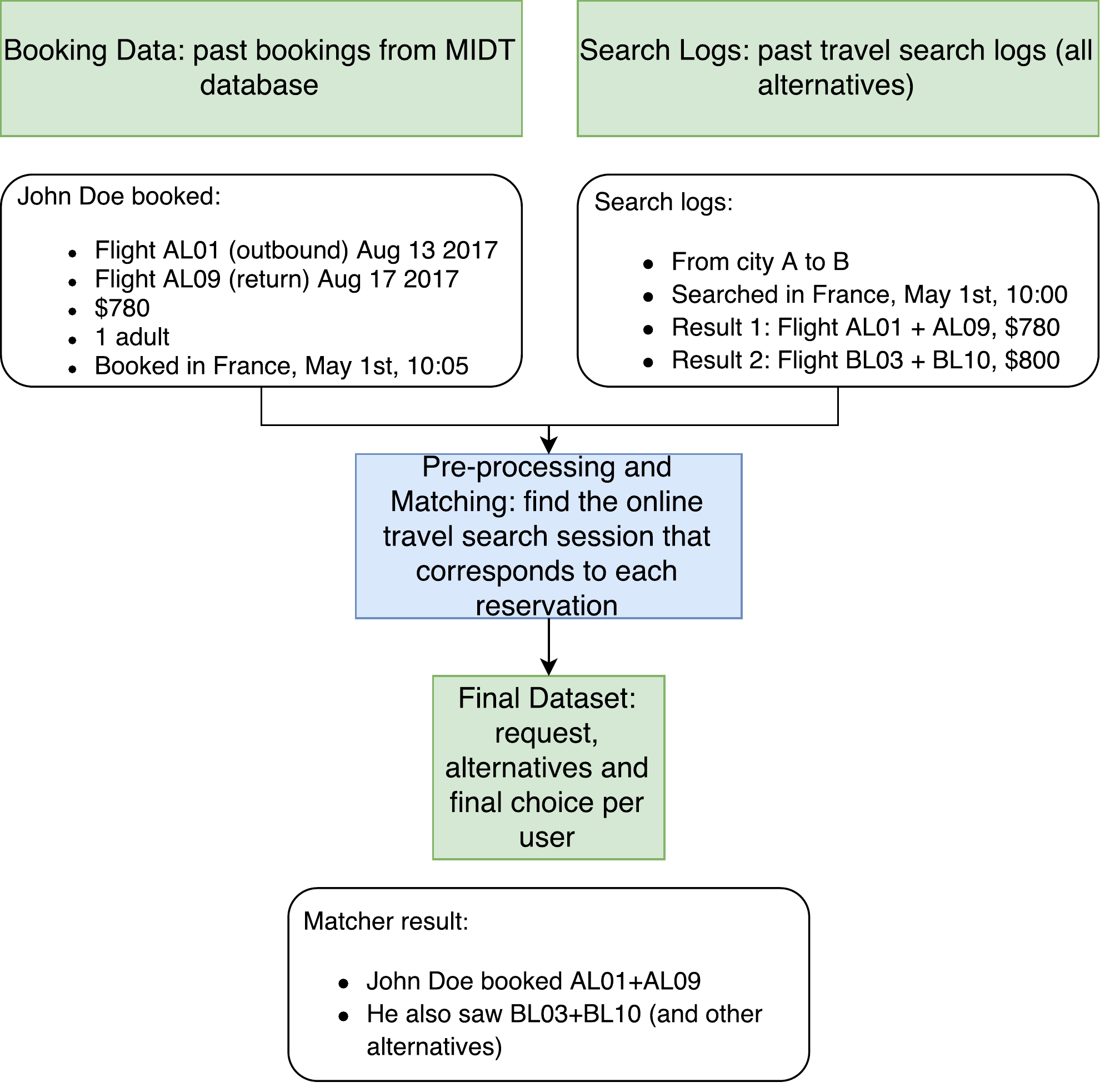}
\caption{Dataset generation through MIDT bookings and search log matching. }
\label{fig:matching}
\end{figure}

The matching process itself is challenging due to the high volume of data (i.e., around 100 GB of search logs per days) and to the difference in data sources and formats. We have developed a process to prepare and match these data based on big-data technologies. The process is not perfectly accurate since the the booking and search times differ, and there is no a direct link between these two data sources. The matchings are produced for each booking and search elements using information such as booking and search dates, flight date/number, and origin/destination. The process is summarized in Figure \ref{fig:matching}.

For this study, we have only considered certain airlines and medium-haul markets. Note that travelers can behave differently on different markets.
For example, the relative values of travel time and price are likely to be different on a long-haul market than a short-haul one. In addition, the data is restricted to travel requests concerning round trips. 

The resulting dataset\footnote{Data and code available for download here: \url{https://amadeus.box.com/s/uv5ctxle5u5p1pysh5kiofgf4s88mxks}\label{footnote}} contains both numerical and categorical features, such as price of ticket, number of connecting flights and airline. The complete list of used features is presented in Table \ref{tbl:featPNR}. Certain features are shared between all of the proposed itineraries per user (e.g., origin) while others change per itinerary and per user (e.g., price). In total, there are 33951 unique users in the dataset, which are divided into training, validation and test set for the experiments. Note that although potentially useful, for legal reasons no personal features (such as gender or age) were used in this work. In addition, the only performed feature engineering is to transform departure/arrival time into circular coordinates.


We have compared our method against the classic MNL and an alternative machine learning method (which we will call ML for simplicity). The ML method consists of training a classifier (gradient boosting tree in our case) on all sessions grouped together and shuffled. The classifier will thus try to learn if an alternative was chosen or not by some user.  Finally, each user session is regrouped and the probability estimates of each choice normalized using the softmax function.

The proposed deep choice model was implemented with Tensorflow and is available for download\textsuperscript{\ref{footnote}}. Moreover, we have used the MNL model as implemented by the Larch  open toolbox \cite{Larch}. Finally, the ML method was implemented using Scikit-learn \cite{sklearn} and XGBoost \cite{xgb}.




Due to computational constraints, we have not performed an extensive hyper parameter tuning for our deep choice model. The used parameters are detailed in Table \ref{tbl:params}. In the case of ML, a random search with 3-fold cross validation was conducted. The MNL algorithm has no tunable parameters.

\begin{table}
  \caption{Type (numerical or categorical) and range/cardinality of each feature  used to represent the itineraries of the sessions}
  \label{tbl:featPNR}
  \begin{tabular}{ccc}
    \toprule
    Feature & Type & Range/Card. \\
    \midrule
Origin/Destination & Categorical &  97\\ 
Search Office & Categorical & 11\\ 
Airline (of first flight)& Categorical & 63\\
Stay Saturday & Binary & $\{$0,1$\}$\\  
Continental Trip & Binary & $\{$0,1$\}$\\  
Domestic Trip & Binary & $\{$0,1$\}$\\  
Price (EUR) & Numerical & [77,16780]\\  
Stay duration (minutes)& Numerical & [120,434000] \\  
Trip duration (minutes) & Numerical & [105, 4314]\\  
Number connections & Numerical & [2,6]\\  
Number airlines & Numerical & [1,4]\\  
Days to departure & Numerical & [0, 343]\\  
Departure weekday & Numerical & [0,6]\\  
Outbound departure time,  & TimeDate & [00:00, 23:59]\\  
Outbound arrival time,  & TimeDate & [00:00, 23:59] \\  
  \bottomrule
\end{tabular}
\end{table}

\begin{table}
  \caption{Hyper-parameters used for the deep choice model}
  \label{tbl:params}
  \begin{tabular}{cc}
    \toprule
    Name & Value\\
    \midrule
Opt. algorithm & Adagrad\\  
Learning rate  &  0.1 \\  
Batch size & 128 \\
Memory size & 128 \\  
N. Layers Enc. & 1 \\  
Cell type & LSTM  \\  
Grad. Clipping  & 8.0  \\ 
k & 5  \\
  \bottomrule
\end{tabular}
\end{table}

\subsection{Results}

The three models  are evaluated using top-N accuracy and other business-centric metrics. We compare our approach to the MNL and ML models, as well as with two simple rule-based methods: the predicted choice is the first alternative of each choice set, and the predicted choice is the alternative presenting the shortest total flight time. In case of a tie, the first alternative fulfilling the condition is chosen. It should be noted that the alternatives are sorted by ascending price, but multiple alternatives can have equal price and flight time.

As seen in Table~\ref{tbl:resAcc}, our approach outperforms all others in terms of top-1 and top-5 accuracy. 
It should be noted that for applications such as dynamic pricing, a small difference in top-1 and top-5 prediction accuracy can lead to a significant increase in profit. For example, if an airline knows that their itinerary is the most likely choice of a user, they can increase the price slightly. Even a one percent increase per user can lead to a significant increase in overall profit \cite{paperRodrigo}.

\begin{figure}
\centering
\includegraphics[scale=0.48]{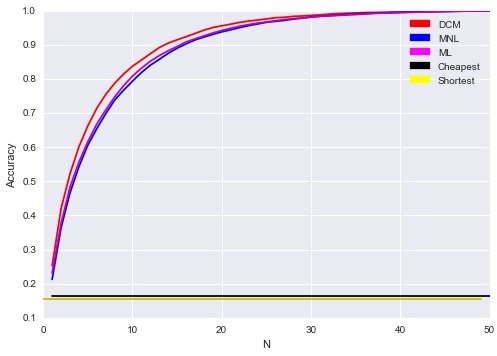}
\caption{Top-N accuracy for the compared methods. }
\label{fig:gracia_cumChoices}
\end{figure}

Figure~\ref{fig:gracia_cumChoices} shows the top-N accuracy for all compared methods. We can appreciate that the difference in accuracy is greater as more alternatives are considered in the computation, the maximum being within the top 15 alternatives. This is of particular importance for ranking the results of a flight search since most websites show approximately 15 results per page,  and users usually look at the first page in more detail.

\begin{figure}
\centering
\includegraphics[scale=0.48]{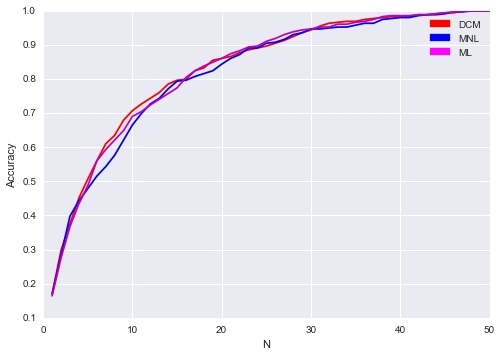}
\caption{Top-N accuracy for the compared methods. Evaluation on a subset of the dataset consisting of a single origin/destination pair. }
\label{fig:gracia_cumChoices_1OD}
\end{figure}

In addition, we have also calculated the top-N accuracy on a reduced subset of the dataset containing only one origin/destination pair. This smaller dataset only contains 1617 users. As we can see in Figure~\ref{fig:gracia_cumChoices_1OD}, all methods perform similarly, although our method still obtains better top-1 and top-5 accuracies. This shows that on pre-segmented dataset, the MNL model is able to perform approximately as well as other more complex methods. Nevertheless, having to pre-segment the dataset and generate one model per segment presents several challenges, which are avoided with our method.

Moreover, we calculate the percentage of sessions that have the real choice in the top 15 alternatives but predicted choice after the top 15 for each of the methods (see Table~\ref{tbl:res_lossCR}). Results show that our method produces less errors with respect to this metric, which has a significant business importance given that not placing the optimal alternative in the first page of the search results could lead to a lower conversion rate.

\begin{figure}
\centering
\includegraphics[scale=0.43]{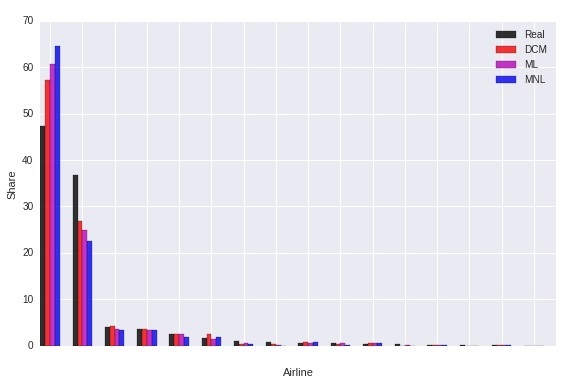}
\caption{Real and predicted airline share for the compared methods. The airlines have been anonymized.}
\label{fig:airlineShares}
\end{figure}

Finally, we calculate the global real and predicted airline market shares. The market shares are calculated by counting the number of real and predicted choices associated to each airline (hard prediction), and  normalizing by the number of sessions in the dataset.  Results are presented in  Figure \ref{fig:airlineShares}. One can notice that our method better approximates the real market share per airline. A good estimation of the market shares is of great importance for different airline applications such as schedule planning and the prediction of the potential impact of a new flight/route.





\begin{table}
  \caption{Top-1 and top-5 accuracy for the compared models}
  \label{tbl:resAcc}
  \begin{tabular}{ccc}
    \toprule
    Method &Top-1 acc. &Top-5 acc.\\
    \midrule
DCM & 25.3 & 66.3\\  
ML & 23.1  & 61.7\\  
MNL & 21.2 & 60.6\\  
Cheapest & 16.4 & 16.4\\  
Shortest & 15.4 & 15.4\\
  \bottomrule
\end{tabular}
\end{table}

\begin{table}
  \caption{Percentage of sessions that have the real choice within the top 15 alternatives but predicted choice after the top 15}
  \label{tbl:res_lossCR}
  \begin{tabular}{cc}
    \toprule
    Method &Percentage \\
    \midrule
DCM & 6.9\\  
MNL & 7.1 \\  
ML &  13.6 \\  
  \bottomrule
\end{tabular}
\end{table}

\section{Conclusions and Future Work}
\label{sec:conclusions}

In this work we propose a new deep choice model based on Pointer Networks, a recent neural architecture that combines Recurrent Neural Networks with the Attention Mechanism to point to elements in an input sequence. This approach specifically targets problems where the outputs are discrete and correspond to positions in the input. In the context of choice modeling, given an input sequence of alternatives presented to a user, the model predicts the one that is going to be selected.


The proposed model was evaluated on a real dataset of  matched airline bookings and online search logs. The data contains searches and
bookings on a set of European origin/destination markets and airlines. The performance of our method was compared against the one obtained by the classic Multinomial Logit model and a gradient boosting tree based method. Results show that the proposed method is able to outperform both models in terms of prediction accuracy and additional business metrics. Moreover, our model presents several advantages over the traditional MNL approach: non-linearity with respect to the input features, no statistical  independence assumptions of the alternatives, and no previous data segmentation is required. In addition, the use of RNN allows the model to take into account the order of the alternatives.



In the future, it would be interesting to measure to which extent does the order in which the choices are input into the model alters the results.  Furthermore,  we would like to determine how we could use our model to gain the same types of insights  that can be obtained with the MNL model. For example, since MNL is linear, we can directly use the weights associated to each feature to compute business metrics such as the elasticity of the revenue with respect to the ticket price or trip duration.  

From a business perspective, we intend to test if the model could be used  for price optimization, and if the prediction accuracy improvement results in a real increase in airline ticket sales and overall profit. Finally, we will test this approach on a bigger scale and implement it at an industrial scale.

\begin{acks}
The authors would like to thank their colleague Alix Lheritier and Jan Margeta for providing valuable input and discussion during  the paper. We would also like to thank our colleague Eoin Thomas for  proofreading the manuscript. 
\end{acks}

\newpage
\bibliographystyle{ACM-Reference-Format}
\bibliography{yo}

\end{document}